\documentclass[authoryear]{elsarticle}
\usepackage{graphicx}
\usepackage{caption}
\usepackage{tikz}
\usepackage{bbm}
\usepackage{amsmath}
\DeclareMathOperator*{\argmax}{arg\,max}
\graphicspath{{./figs/}}
\newcommand\inputpgf[2]{{
    \let\pgfimageWithoutPath\pgfimage
    \renewcommand{\pgfimage}[2][]{\pgfimageWithoutPath[##1]{#1/##2}}
    \input{#1/#2}
}}

\begin{document}
\date{}

\title{ExGate: Externally Controlled Gating for Feature-based Attention in Artificial Neural Networks}

\author{J. D. Son\corref{cor1}}
\ead{snxjar002@myuct.ac.za}

\author{A. K. Mishra}
\ead{amit.mishra@uct.ac.za}

\cortext[cor1]{Corresponding Author}
\address{Department of Electrical Engineering, University of Cape Town, Menzies Building, Rondebosch, Cape Town, 7700, South Africa}
\begin{abstract}
    Perceptual capabilities of artificial systems have come a long way since the advent of deep learning.
    These methods have proven to be effective, however they are not as efficient as their biological counterparts.
    Visual attention is a set of mechanisms that are employed in biological visual systems to ease computational load by only processing pertinent parts of the stimuli.
    This paper addresses the implementation of top-down, feature-based attention in an artificial neural network by use of externally controlled neuron gating.
    Our results showed a 5\% increase in classification accuracy on the CIFAR-10 dataset versus a non-gated version, while adding very few parameters.
    Our gated model also produces more reasonable errors in predictions by drastically reducing prediction of classes that belong to a different category to the true class.
\end{abstract}

\maketitle
\section{Introduction}
Artificially intelligent agents are often designed with bottom-up processing mind. 
Agents perceive the world through sensors, make decisions based on the sensory information and then perform an action based on the decisions.
It is often the case that agents are faced with interpreting large amounts of sensory data.
Processing all of the incoming data can be too computationally expensive for constrained systems such as mobile robots.

Advances in machine learning, notably with the rise of deep learning \citep{lecun2015deep}, have made a significant contribution to computer vision by taking ideas from biological vision to develop better artificial neural network architectures \citep{szegedy2015going}. 
The majority of these architectures follow the basic structure of the convolutional neural networks (CNNs) introduced in \citep{lecun1989backpropagation, lecun1995convolutional}.
For many years CNN model progression has generally consisted of taking that basic structure but adding more layers to make the networks deeper \citep{krizhevsky2012imagenet, simonyan2014very}.
The complexity of these models has restricted their usage to high-performance computers that are not suitable for constrained systems.
The need to create less demanding models has not gone unnoticed and research has focussed on developing alternative structures to improve efficiency and performance \citep{szegedy2015going, he2016deep, szegedy2017inception}.
In even more extreme cases performance has been somewhat sacrificed to create far less demanding models \citep{iandola2016squeezenet, howard2017mobilenets, zhang1707shufflenet}. 

These traditional CNN based architectures have been very effective at various tasks involving visual data, however the trained networks are static.
This means that the same transformations will always be applied to all of the input data.
There is no option to modify the behaviour of the network in real-time, unlike what humans are capable of.

Humans have computationally strained resources, yet we are capable of interpreting enormous amounts of sensory data with ease.
This is possible because our brains utilize various mechanisms to select portions of the incoming data to operate on and we can also constrain our interpretations of that data as well \citep{summerfield2009expectation}.
These mechanisms have been studied extensively in fields such as neuroscience and cognitive science, especially with regards to how we, and other animals, process visual stimuli so efficiently \citep{desimone1995neural, yantis2003cortical, summerfield2009expectation, carrasco2011visual, moore2017neural}.
Visual attention is one of the most studied mechanisms for modulation of visual perception.
This paper focusses on top-down feature-based attention (see Section \ref{sec:feat-att}) to modulate the behaviour of a feedforward neural network in an image classification task.

Further background to visual attention is provided in Section \ref{sec:vis-att} to clarify the connection between this research for artificial systems with the research on biological systems.
The background supplements the development of our novel mechanism for feature-based attention presented in Section \ref{sec:gating}.
The experimental procedure is discussed in Section \ref{sec:experiments} and the results are presented in Section \ref{sec:results} along with the discussion of the results.
Lastly we draw conclusions in Section \ref{sec:conclusion}.

\section{Visual Attention} \label{sec:vis-att}
Visual attention is one of the mechanisms that our brains use to select pertinent aspects of visual stimuli.
It can manifest itself as a bottom-up process, where portions of the stimuli are attended to automatically based on salient features \citep{desimone1995neural}.
Attention can also be controlled in a top-down manner, where a current behavioural state may favour certain aspects of the stimuli \citep{gilbert2013top, noudoost2010top}.
Our research here focusses on top-down attention, as we wish to find ways to control aspects of low-level perceptual systems from higher-level cognitive agents.

There are multiple ways in which attention can be controlled in a top-down manner.
Attention can be directed to a specific location (spatial attention), to certain features, and to distinct objects \citep{noudoost2010top, gilbert2013top, yantis2003cortical}.
Spatial attention has been used to great effect in recent deep learning works such as \citep{xu2015show, wu2016encode, lu2017knowing}.
These networks introduce dynamic capabilities by using recurrent neural networks (RNNs) to supply contextual information based on the history of inputs.
This allows the models to learn how the sequential structure of sentences in the training data relates to specific locations of an image that should be attended to when generating new captions.
The feedback inherent in the RNNs make this a form of ``top-down" attention, where the RNN is acting as a primitive memory system that guides spatial attention.

\subsection{Feature-based Visual Attention} \label{sec:feat-att}
Our work focusses on feature-based attention and we aim to introduce a greater level of top-down control than the works mentioned above.
Feature-based attention modulates neuron activity based on the relationship between features of interest and the stimulus \citep{moore2017neural}.
Unlike spatial attention which attends to specific locations of a stimulus, feature-based attention tries to measure how the feature components of stimulus compares to the features it believes are relevant to the task at hand.
\citep{maunsell2006feature} show that certain neurons in the visual cortex have enhanced responses to stimuli relevant to the agent's behavioural goals. 
\citep{gilbert2013top} also suggests that top-down control can engage relevant components of stimuli while also discarding irrelevant components. 

Such a mechanism could also be used to improve the performance of artificial neural networks.
Suppressing irrelevant features and enhancing relevant ones should increase the signal-to-noise ratio and should therefore be easier for a classifier to discriminate between classes.
Instead of applying the exact same transformations on input data, we can find a way to modulate the behaviour of the neural network depending on the goals or desires of the agent.

\section{Externally Controlled Feature Gating} \label{sec:gating}
This section presents a gating mechanism that can be controlled symbolically, rather than through neural connectivity.   
Such an approach was chosen to facilitate the greater vision of a neuro-symbolic system where symbolic cognitive architectures can influence neural processing in a top-down manner.
This implementation differs from the approach taken in common deep learning models that use attention mechanisms such as \citep{bahdanau2014neural, xu2015show, wu2016encode}.
In these cases the attention model is parameterised by an additional feedforward neural network.
These models use hidden states from recurrent neural networks as an input to the attention model, whereas in our case it is assumed that a symbolic oracle generates the top-down signals.

The type of visual attention used in \citep{xu2015show, lu2017knowing} only accounts for spatial attention.
Their models focus visual attention on specific locations within the visual field, but do not attend to specific features.
Neurons in the visual cortex can be modulated to enhance certain visual features based on the current behavioural preference \citep{maunsell2006feature}.
We propose a mechanism that multiplexes different gating layers depending on a specific task. 
This method allows us to train one set of feature detecting neurons, but by gating their outputs we can attend to specific neuron populations based on the current task.  

\subsection{Gating Units}
A gating unit chooses whether to pass or suppress its input based on trainable parameters.
The gating units used in our architecture act independently of the input, and therefore only require training of a bias value.

The output, $g$,  of a single gating unit can be expressed as

\begin{equation} \label{eqn:gate_output}
    g = a \cdot \sigma (b)
\end{equation}

Where $a$ is the activation of the neuron that is being gated, $\sigma (.)$ is the sigmoid function, and $b$ is the trainable bias. 
This mechanism allows us to modulate neuron activations by choosing different sets of biases, rather than modifying the weights that determine how neurons respond to input stimuli.

Eq. \ref{eqn:gate_output} can be extended to the case where there are multiple neurons, layers and tasks.
This results in a gating vector $\mathbf{g^{l}}$ given by Eq. \ref{eqn:gate_vector}.

\begin{equation} \label{eqn:gate_vector}
    \mathbf{g}^{l} = 
        \begin{bmatrix}
            a^{l}_{0, 0} \cdot \sigma (b^{l}_{0,0}) \\
            \cdot \\
            \cdot \\
            \cdot \\
            a^{l}_{m,t} \cdot \sigma (b^{l}_{m,t})
        \end{bmatrix}
\end{equation}

where $l$ refers to the layer number, $m$ refers to the index of the neuron and $t$ refers to the index of the task the gating layer belongs to.

\subsection{Architecture}
\begin{figure}[ht] 
    \centering
        \includegraphics[width=0.6\textwidth]{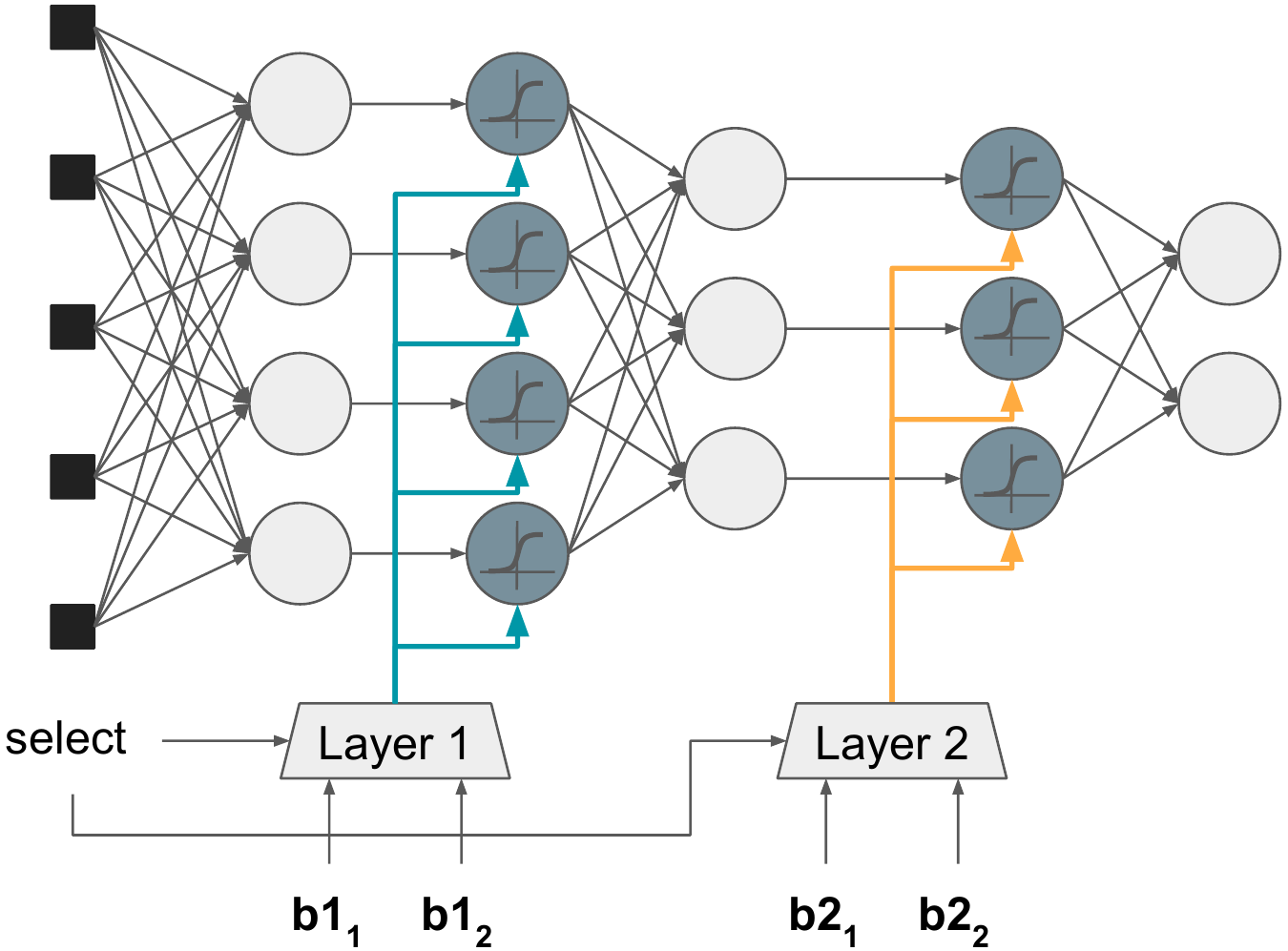}
        \caption{The basic architecture consists of a feedforward artificial neural network that has trainable gating layers after each hidden layer.
                 The "select" input is a symbolic input that determines which set of gate biases to select.
                 Where $\mathbf{bl_n}$ refers to the bias vector at layer, $l$, and task, $t$.} \label{fig:architecture}
\end{figure}

An example of how these gating layers can be incorporated into a feedforward neural network is shown in Figure \ref{fig:architecture}.
The gating layers are trained independently so that they become specialised on specific types of input data.

This approach is less computationally expensive than adding additional fully connected layers as is often the case with other multi-task learning approaches \citep{ruder2017overview}.
For the entire network we are only adding $T \cdot N_{h}$ number of parameters, where $T$ is the total number of tasks and $N_{h}$ is the total number of hidden neurons. 
Our approach is also closer to biological findings where individual neurons have shown the ability to switch their behaviour based on a visual categorization task \citep{cromer2010representation}.

Keeping the neuron weights the same despite changing the task is commonly used in deep learning \citep{yosinski2014transferable, ruder2017overview}, notably for visual data where the basic representations of features is common regardless of the task.
Our work is novel in that we are able to modify which of these common visual features are most relevant to the current goal, rather than fine-tuning a classification stage after the input features are extracted.

\subsection{Problem Details}
For our experiments here we will treat classifying different categories of images as different tasks.
Let $\mathcal{X}$ be the input space and $\mathcal{Y}$ be the output space of our supervised learning problem.
Let $\mathcal{D} = \{(\mathbf{x}_{i}, y_{i}) | i = 1:N\}$ be a given dataset where $\mathbf{x}_{i} \in \mathcal{X}$ are sample inputs, and $y_{i} \in \mathcal{Y}$ are the corresponding labels.
Let $\mathcal{C} = \{c_{1}, ..., c_{M}\}$ be a set of defined categories that labels in $\mathcal{Y}$ can belong to, where $M$ is the total number of categories.
For now we are assuming that the labels can be grouped into categories by an oracle (a human in this case). 
We also only consider the case that each label belongs to a single category.

We wish to create a feedforward, fully-connected artificial neural network that produces the output given by $z_{i} = f_{\theta_{c}}(x_{i})$, where $x_{i}$ is the $i$-th input pattern, and $\theta_{c}$ are the trainable parameters for the $c^{th}$ category, $c \in \mathcal{C}$. 
The predicted labels from the network are then given by $\hat{y}_i = \argmax_{x_{i} \in \mathcal{X}} f_{\theta_{c}}(x_i) $
Let $Y^{(c)}$ be a subset of $\mathcal{Y}$ containing elements belonging to category $c, c \in \mathcal{C}$. 
The parameters for each category, ${\theta_{c}}$, are trained using gradient descent methods on training data, $\mathcal{D}_{train, c} = \{(\mathbf{x}_{i}, y_{i}) | i = 1:N_{c}, y_{i} \in Y^{(c)}\}$ where $N_{c}$ is the number of labelled samples in category $c, c \in \mathcal{C}$.
The objective of training is to find the parameters that minimize some loss function $L(y_{i}, \hat{y}_{i})$.

\subsection{Evaluation Metrics}
The standard classification metrics of accuracy and loss can be used to compare the performance of model with and without feature gating.
These metrics do not provide any insight into the what effect the gating may have had on the base model neurons to impact performance.
We hypothesise that the gating should minimize predictions across different categories because they select a subset of features that are only relevant to the ``cued" task.
Given some test data $\mathcal{D}_{test} = \{(\mathbf{x}_{i}, y_{i})|i = 1:N_{test})\}$, where $N_{test}$ is the total number of test samples, we can define $\hat{Y}^{(c)}$ as the set of predicted labels from the test data with category, $c$, inferred from the true label, $y_{i}$. 
Based on our hypothesis we declare a new metric in Eq. \ref{eqn:cat_isolation} to quantify the effect of feature gating on class prediction based on categorical membership.

\begin{equation} \label{eqn:cat_isolation}
    \phi = \frac{1}{N_{test}} \sum\limits_{c \in \mathcal{C}} \sum\limits_{\hat{y} \in \hat{Y}^{(c)}} {[ \hat{y} \in Y^{(c)} ]}
\end{equation}

Where the $[.]$ are Iverson brackets, 
\[
    [P]=\begin{cases}
     1 & \text{if P is true} \\ 
     0 & \text{otherwise}
    \end{cases}
\]
We term this metric ``categorical isolation'' and is defined as being the percentage of predicted labels belonging to the same category as the true label.
The higher the categorical isolation the better the model is at discriminating classes between different categories.

\section{Methods} \label{sec:experiments}
The capabilities of our architecture were tested on an image classification task using the CIFAR10 dataset \citep{krizhevsky2014cifar}.
PyTorch was used as the framework, as it allows for dynamic computational graphs which are required for switching between gating layers during inference.
The hypothesis is that overall classification accuracy can be improved by switching between specialized gating layers that are trained on categorized subsets of the original data.

\subsection{Dataset}
CIFAR10 consists of thousands of $32 \times 32$ images containing unique instances of 10 different classes.
The class labels are airplane, automobile, bird, cat, deer, dog, frog, horse, ship and truck. 
The default training set consists of 50000 images and the default test set consists of 10000 images.
Each class has exactly the same number of samples in both the training and test data.

\begin{figure} 
    \begin{center}
        \includegraphics[width=\textwidth]{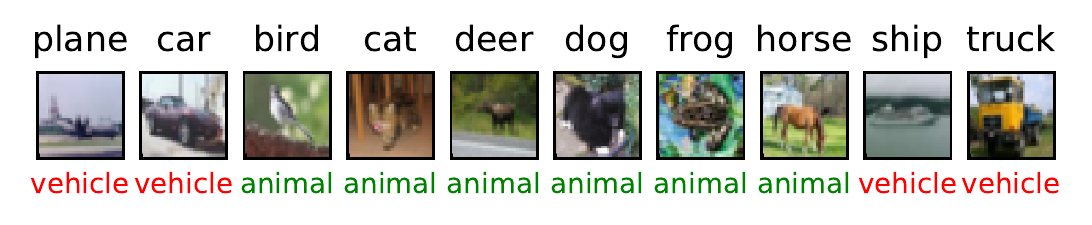}
    \end{center} 
    \caption{Randomly selected samples from each class. The title above each image indicates the label for the class and the title below indicates the category the class belongs to.} \label{fig:samples}
\end{figure}

The dataset was divided into two broad categories: ``vehicles", and ``animals".
These categories acted as two separate "tasks" for the network to perform. 
Samples of the images and categories are provided in Figure \ref{fig:samples}. 

Each set of gating layers was trained and tested on a subset of the data based on these categories.
One set of gating layers was trained on only vehicles and the other set was trained only on animals.
This ensures that the gates are specialised to a specific category.
During training $10\%$ of the original training data were used as a validation set. 

The images were converted to grayscale and normalized. 
No other pre-processing was performed, as we are only looking for a comparative improvement between gated and non-gated versions of the neural network.

\subsection{Neural Network Parameters}
The base model was a fully-connected feedforward neural network with the two hidden layers.
The first hidden layer had 256 neurons and the second hidden layer had 128 neurons.
Neurons in the hidden layers had a Rectified Linear Unit (ReLU) type activation function. 
Dropout was applied to the base model during training with a dropout rate of 50\%.
There were 10 linear output neurons (one per class) with a log-softmax function applied to the output to generate a log-probability output.

The gated version of this network had gating layers, as specified in Section \ref{sec:gating}, after each hidden layer.
The biases in the gated layers were initialized to 0.

\subsection{Training and Testing Procedure}
Trainable parameters were updated after calculating the average loss for mini-batches consisting of 32 randomly sampled images from the training set.
Validation was performed using the validation set after a set number of updates to monitor for overfitting on the data used for calculating the gradients.
The negative log-likelihood loss was used as the loss function to optimize, using the RMSprop optimizer with a learning rate of $1e^{{-2}}$.
The set of parameters that produced the lowest validation loss were saved and used to test the model on the holdout test data.

Before training the gates, the base model was trained on all of the training data to learn the feature representations generic to all classes.
When training the gated versions the dataset was sampled such that each set of gates was presented with classes exclusive to the category it was being trained for.
For example, when training the ``vehicle'' gates, only images of planes, cars, ships and trucks were presented to the network.
The parameters for non-gating layers were initialized using the best parameters from the base model, and were not modified further during training.
This ensures that the same neurons are being gated for both categories rather than having different sets of neuron parameters for each.
We are only interested in learning the category dependant gate biases.
As with the base model, the set of gates producing the lowest validation loss for each category were saved and used in the final testing procedure.

During testing the model is cued by using categories inferred directly from the image labels e.g. if the label was "truck" the category would be "vehicles".
For every image presented to the model the category was used to select which set of gates to use for inference, as illustrated in Figure \ref{fig:architecture}.
In this paper we are only evaluating the effectiveness of the gating mechanism and not how an agent automatically selects a category.

\section{Results and Discussions} \label{sec:results}
\begin{figure}
    \begin{center}
        \includegraphics[width=0.7\textwidth]{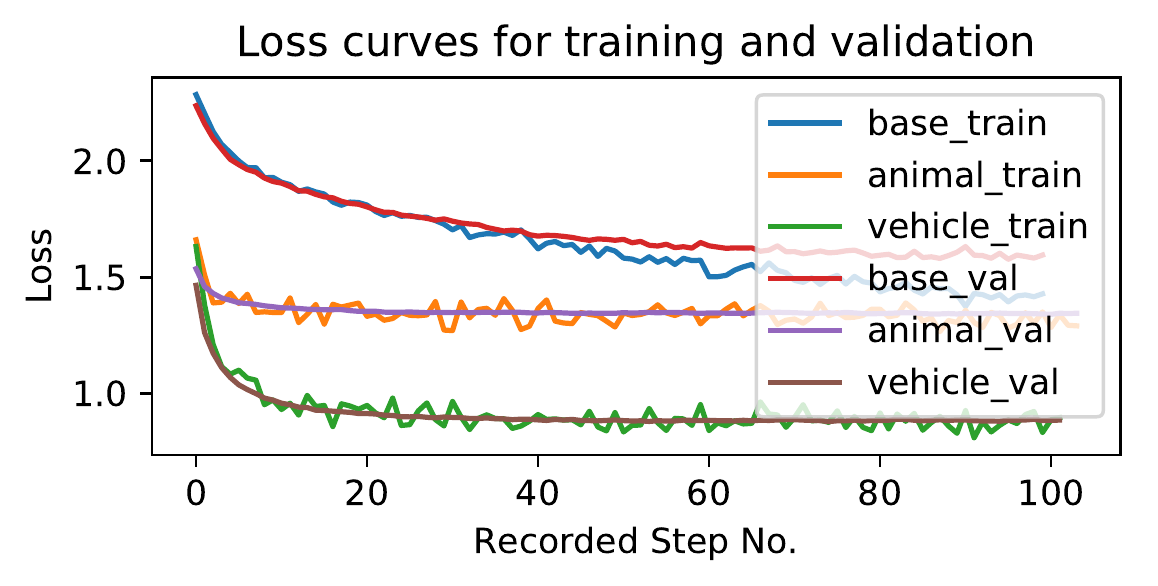}
    \end{center} 
    \caption{The loss curves are plotted based on a fixed number of recorded steps, not number of iterations. ``Base" refers to the model without gating, ``animal" and ``vehicle'' refers to the gated models trained on classes belonging to the animal and vehicle categories respectively. ``Train" indicates the loss for training data and ``val" indicates the loss for validation data.} \label{fig:loss}
\end{figure}

The results of the experiments are presented in this section along with relevant discussions.
The loss curves during the training phase are shown in Figure \ref{fig:loss}.

The gated versions show much lower validation losses than the base model, which indicates they are improving upon the non-gated model.
The fact that the gated models rely on pre-trained neuron weights and biases makes training the gating biases quick to train especially because there are so few parameters in comparison to the number of parameters needed for even one neuron.
This is beneficial because one can reuse the same base model and very quickly train different gating layers. 
This is likely even quicker than the fine-tuning techniques used in transfer learning for deep learning models, where the final classification layer is often a fully connected layer that is tuned to different data \citep{yosinski2014transferable}.

Table \ref{tbl:eval} shows the performance of the base model with and without feature gating evaluated on the holdout test data.
It compares the test loss, accuracy and categorical isolation as defined in Eq. \ref{eqn:cat_isolation}.

\begin{table}[ht]
    \centering
    \caption{Model Evaluation on Test Data} \label{tbl:eval}
    \begin{tabular}{ l | c c c}  
        \hline
        \bf Model & \bf Test Loss & \bf Test Accuracy & \bf Categorical Isolation\\
        \hline
        Non-gated & 1.58 & 44.9\% & 83.0\%\\
        Gated & \bf 1.33 & \bf 50.0\% & \bf 98.2\%\\
    \end{tabular}
\end{table}

The results show that the feature gating reduces the test loss by 0.25, increases the total classification accuracy by 5.1\%, and increases categorical isolation by 15.2\%.
This represents a strong improvement on performance in terms of accuracy, while adding only few trainable parameters.
Even a single additional hidden neuron would result in more parameters than a full gating layer.
The improvement in terms of categorical isolation is significant, and it shows that features can have a drastic difference in importance depending on the active behavioural goals.
This is an import result because it means the network is producing outputs that are more reasonable considering the context of the situation.

\begin{figure} 
    \begin{center}
        \includegraphics[width=0.9\textwidth]{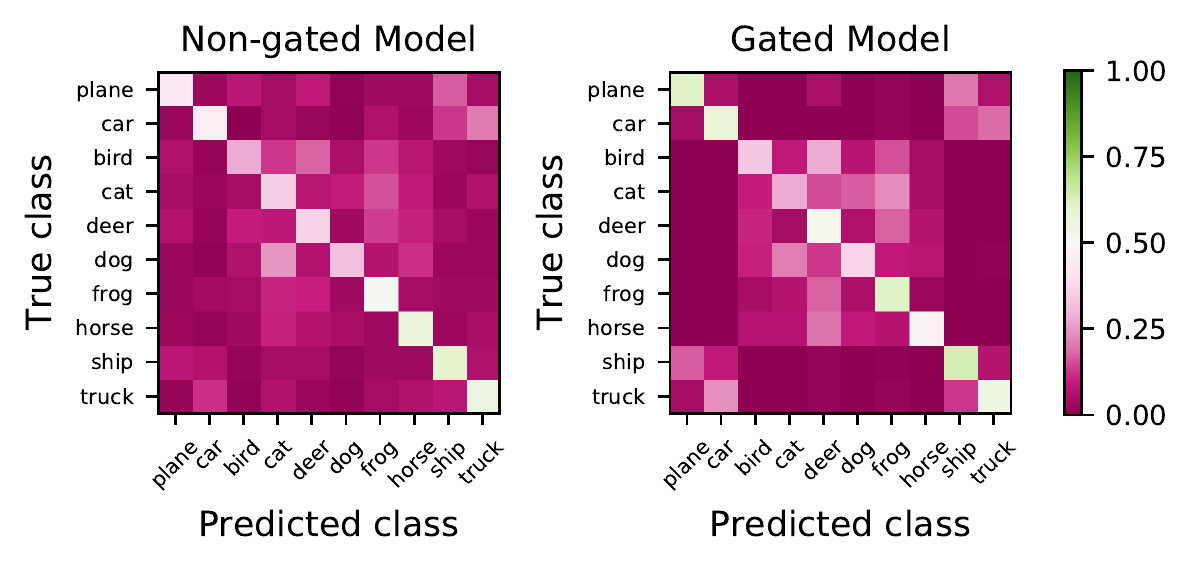}
    \end{center} 
    \caption{The confusion matrices show what percentage of ground truth classes were classified as by the model during testing. The shown confusion matrices were normalized such that each element lies between 0 and 1. \textbf{left:} base model with no feature gating applied, \textbf{right:} base model with feature gating applied.} \label{fig:cm_compare}
\end{figure}

The categorical isolation metric is useful for comparing between different models, but does not provide all the details of the networks behaviour.
It could be possible that the network favours specific categories, which would result in a high categorical isolation, but it might not be producing reasonable outputs for other categories.
To further assist in interpreting the behaviour of the network we direct the readers attention to Figure \ref{fig:cm_compare}.

Of particular interest here is the results of classification between the two different categories of classes.
In the non-gated model there are a number of instances where the model predicted classes from one category as classes from the other category.
In contrast the dark purple regions of the gated model coincide with regions where the predicted classes do not belong to the same category as the true class.
This shows that the gated model almost completely nullifies all cases of misclassification between categories.
The feature gating worked as expected to discriminate classes based on the features relevant to the current goal.

\begin{figure} 
    \begin{center}
        \includegraphics{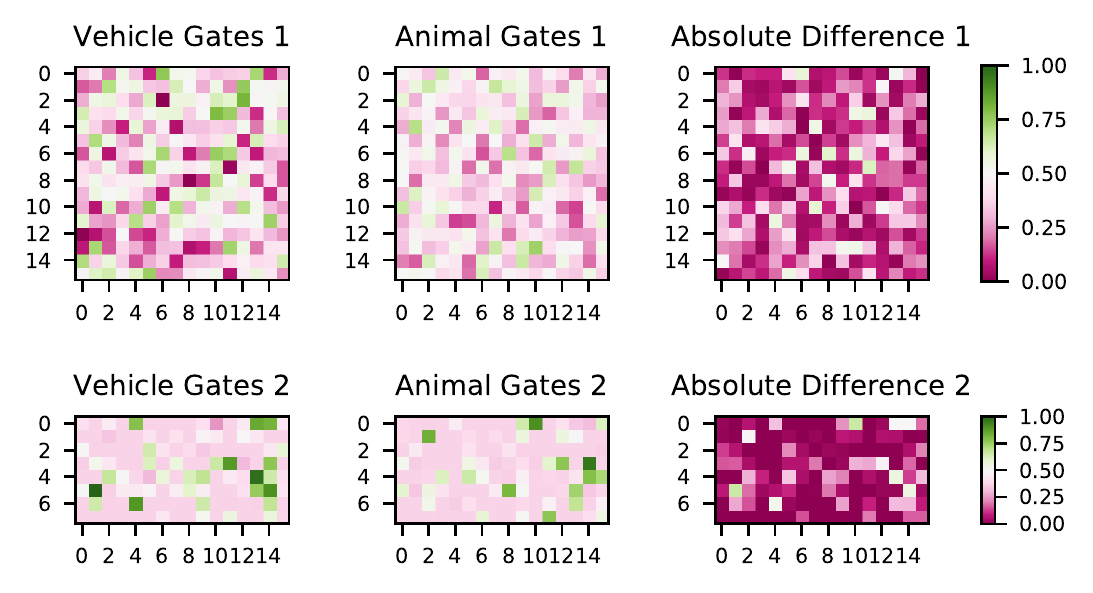}
        \caption{A comparison of the gate bias vectors visualized as 2-D images. The first column shows the gates used for the `vehicle' category,
                 the second column shows the gates used for the `animals' category, and the last column shows the absolute difference between the gates used for each category.
                 The values are normalized to allow for a better visual comparison between the different layers.} \label{fig:gates}
    \end{center}
\end{figure}

We also wish to explore the inner workings of the gating and not just examine the overall effect.
The gate bias vectors are visualized as 2-D images in Figure \ref{fig:gates}. to illustrate the behavioural changes of individual neurons based on the categories.
This may provide useful insights into how gating influences the network at a neuronal level.

The absolute difference between the gates in the first layer shows that a large number of individual biases are different between categories.
These bias changes indicate that neurons in the first layer are highly sensitive to changes in task.
In contrast the second layer of gates show a larger percentage of the biases remain the same, however there are a few neurons that exhibit large changes between categories.
This is an indication that the second layer represents more general features that are present in both categories, but with few specialized neurons that are task dependant.

\begin{figure} 
    \begin{center}
        \includegraphics{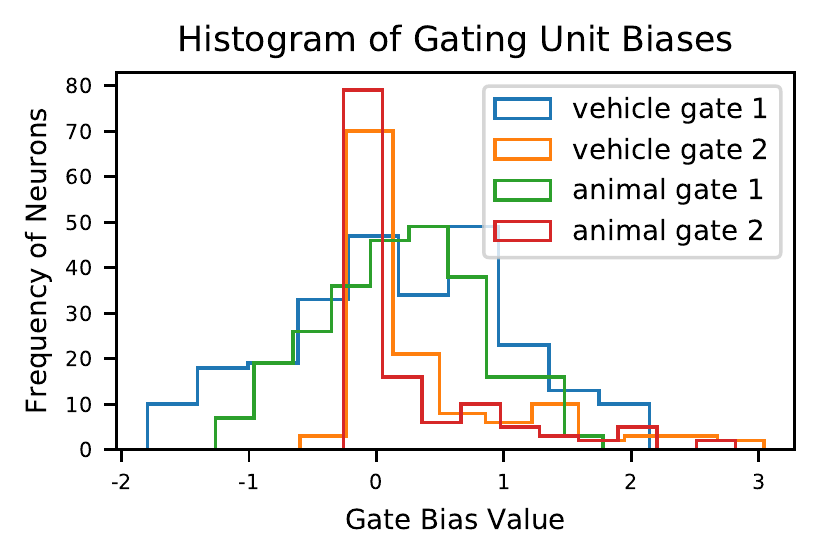}
        \caption{The gate bias histograms show the distribution of the biases to give an idea about the behaviour of the entire population of gating units rather than individual neuron differences.} \label{fig:gates_historgram}
    \end{center}
\end{figure}

We also noticed that the range of the biases are quite distinct between the two layers.
The histogram plots in Figure \ref{fig:gates_historgram}. show how the biases are distributed for each gating layer.
This shows the behaviour of the gating layers as a whole rather than indicating the differences in individual gating units. 

The histogram show that the second layer of gates pass through almost all of neuron activations with 50\% suppression ($\sigma(0) = 0.5$) or more, whereas the first gating layers are more symmetrical about the 50\% suppression point.
The fact that most of the second layer neuron activations are never fully suppressed further supports the idea that the second layer neurons are more general across tasks.
What is also interesting to note is that some of the biases in the second layer have much higher values that in the first layer.
This suggests that some of the neurons in the second layer are highly sensitive to specific features of the input.

\section{Future Work}
The experiment presented in this paper was merely to determine whether or not feature gating in this fashion work and how well it would work.
Further research would include applying our gating mechanism to different datasets with more categories to evaluate how well our solution scales to an increase in the number of categories.
Applying this method to convolutional neural networks (CNNs) is a high priority, as these are more akin to biological visual systems, and are the current state-of-the-art in the majority of visual tasks.
It is expected that our gating mechanism is general enough that it can be applied in the same way to convolutional kernels used in CNNs.

A primary goal for this work is to develop a neuro-symbolic cognitive architecture.
The symbolic system would be better suited to high-level cognitive tasks such as reasoning and problem solving than perceptual tasks where artificial neural networks have been excelling.
Our gating mechanism allows for top-down control of neural-like perception by the symbolic system depending on its goals.
Many other cognitive architectures only consider or implement a bottom-up flow of information that starts by sensing the world, making decisions based on the interpretation of the sensory data, and then acting on these decisions.
They rarely implement the case where the perceptual stage can be influenced in a top-down manner.

This type of neuro-symbolic architecture would be ideally suited for mobile robotics.
Mobile robotics applications such as search and rescue require robots to deal with a variety of situations.
By modifying how a robot perceives the world depending on the situation could result in better perceptual accuracy as shown by our results.

\section{Conclusion} \label{sec:conclusion}
We have implemented a novel approach to feature-based attention in feedforward neural networks.
The approach to use externally controlled, task-dependant gating units shows major improvement in image classification despite only adding a few parameters.
The mechanism is also general enough that it should be possible to replicate in most other forms of artificial neural networks.
Another benefit is that it is designed to allow for easy control from external source, such as a cognitive architecture, to modulate the behaviour of the neural network to enhance perception.
This could lead towards a more tractable form of integrating low-level, neural-like perception with higher-level symbolic cognitive systems.

\section{Acknowledgements}
The authors would like to thank the University of Cape Town for providing the funding for this research.

\bibliographystyle{elsarticle-harv}
\bibliography{bibliography}

\begin{thebibliography}{27}
\expandafter\ifx\csname natexlab\endcsname\relax\def\natexlab#1{#1}\fi
\providecommand{\url}[1]{\texttt{#1}}
\providecommand{\href}[2]{#2}
\providecommand{\path}[1]{#1}
\providecommand{\DOIprefix}{doi:}
\providecommand{\ArXivprefix}{arXiv:}
\providecommand{\URLprefix}{URL: }
\providecommand{\Pubmedprefix}{pmid:}
\providecommand{\doi}[1]{\href{http://dx.doi.org/#1}{\path{#1}}}
\providecommand{\Pubmed}[1]{\href{pmid:#1}{\path{#1}}}
\providecommand{\bibinfo}[2]{#2}
\ifx\xfnm\relax \def\xfnm[#1]{\unskip,\space#1}\fi
\bibitem[{Bahdanau et~al.(2014)Bahdanau, Cho and Bengio}]{bahdanau2014neural}
\bibinfo{author}{Bahdanau, D.}, \bibinfo{author}{Cho, K.},
  \bibinfo{author}{Bengio, Y.}, \bibinfo{year}{2014}.
\newblock \bibinfo{title}{Neural machine translation by jointly learning to
  align and translate}.
\newblock \bibinfo{journal}{arXiv preprint arXiv:1409.0473} .
\bibitem[{Carrasco(2011)}]{carrasco2011visual}
\bibinfo{author}{Carrasco, M.}, \bibinfo{year}{2011}.
\newblock \bibinfo{title}{Visual attention: The past 25 years}.
\newblock \bibinfo{journal}{Vision research} \bibinfo{volume}{51},
  \bibinfo{pages}{1484--1525}.
\bibitem[{Cromer et~al.(2010)Cromer, Roy and Miller}]{cromer2010representation}
\bibinfo{author}{Cromer, J.A.}, \bibinfo{author}{Roy, J.E.},
  \bibinfo{author}{Miller, E.K.}, \bibinfo{year}{2010}.
\newblock \bibinfo{title}{Representation of multiple, independent categories in
  the primate prefrontal cortex}.
\newblock \bibinfo{journal}{Neuron} \bibinfo{volume}{66},
  \bibinfo{pages}{796--807}.
\bibitem[{Desimone and Duncan(1995)}]{desimone1995neural}
\bibinfo{author}{Desimone, R.}, \bibinfo{author}{Duncan, J.},
  \bibinfo{year}{1995}.
\newblock \bibinfo{title}{Neural mechanisms of selective visual attention}.
\newblock \bibinfo{journal}{Annual review of neuroscience}
  \bibinfo{volume}{18}, \bibinfo{pages}{193--222}.
\bibitem[{Gilbert and Li(2013)}]{gilbert2013top}
\bibinfo{author}{Gilbert, C.D.}, \bibinfo{author}{Li, W.},
  \bibinfo{year}{2013}.
\newblock \bibinfo{title}{Top-down influences on visual processing}.
\newblock \bibinfo{journal}{Nature Reviews Neuroscience} \bibinfo{volume}{14},
  \bibinfo{pages}{350}.
\bibitem[{He et~al.(2016)He, Zhang, Ren and Sun}]{he2016deep}
\bibinfo{author}{He, K.}, \bibinfo{author}{Zhang, X.}, \bibinfo{author}{Ren,
  S.}, \bibinfo{author}{Sun, J.}, \bibinfo{year}{2016}.
\newblock \bibinfo{title}{Deep residual learning for image recognition}, in:
  \bibinfo{booktitle}{Proceedings of the IEEE conference on computer vision and
  pattern recognition}, pp. \bibinfo{pages}{770--778}.
\bibitem[{Howard et~al.(2017)Howard, Zhu, Chen, Kalenichenko, Wang, Weyand,
  Andreetto and Adam}]{howard2017mobilenets}
\bibinfo{author}{Howard, A.G.}, \bibinfo{author}{Zhu, M.},
  \bibinfo{author}{Chen, B.}, \bibinfo{author}{Kalenichenko, D.},
  \bibinfo{author}{Wang, W.}, \bibinfo{author}{Weyand, T.},
  \bibinfo{author}{Andreetto, M.}, \bibinfo{author}{Adam, H.},
  \bibinfo{year}{2017}.
\newblock \bibinfo{title}{Mobilenets: Efficient convolutional neural networks
  for mobile vision applications}.
\newblock \bibinfo{journal}{arXiv preprint arXiv:1704.04861} .
\bibitem[{Iandola et~al.(2016)Iandola, Han, Moskewicz, Ashraf, Dally and
  Keutzer}]{iandola2016squeezenet}
\bibinfo{author}{Iandola, F.N.}, \bibinfo{author}{Han, S.},
  \bibinfo{author}{Moskewicz, M.W.}, \bibinfo{author}{Ashraf, K.},
  \bibinfo{author}{Dally, W.J.}, \bibinfo{author}{Keutzer, K.},
  \bibinfo{year}{2016}.
\newblock \bibinfo{title}{Squeezenet: Alexnet-level accuracy with 50x fewer
  parameters and< 0.5 mb model size}.
\newblock \bibinfo{journal}{arXiv preprint arXiv:1602.07360} .
\bibitem[{Krizhevsky et~al.(2014)Krizhevsky, Nair and
  Hinton}]{krizhevsky2014cifar}
\bibinfo{author}{Krizhevsky, A.}, \bibinfo{author}{Nair, V.},
  \bibinfo{author}{Hinton, G.}, \bibinfo{year}{2014}.
\newblock \bibinfo{title}{The cifar-10 dataset}.
\newblock \bibinfo{journal}{online: http://www. cs. toronto. edu/kriz/cifar.
  html} .
\bibitem[{Krizhevsky et~al.(2012)Krizhevsky, Sutskever and
  Hinton}]{krizhevsky2012imagenet}
\bibinfo{author}{Krizhevsky, A.}, \bibinfo{author}{Sutskever, I.},
  \bibinfo{author}{Hinton, G.E.}, \bibinfo{year}{2012}.
\newblock \bibinfo{title}{Imagenet classification with deep convolutional
  neural networks}, in: \bibinfo{booktitle}{Advances in neural information
  processing systems}, pp. \bibinfo{pages}{1097--1105}.
\bibitem[{LeCun et~al.(2015)LeCun, Bengio and Hinton}]{lecun2015deep}
\bibinfo{author}{LeCun, Y.}, \bibinfo{author}{Bengio, Y.},
  \bibinfo{author}{Hinton, G.}, \bibinfo{year}{2015}.
\newblock \bibinfo{title}{Deep learning}.
\newblock \bibinfo{journal}{nature} \bibinfo{volume}{521},
  \bibinfo{pages}{436}.
\bibitem[{LeCun et~al.(1995)LeCun, Bengio et~al.}]{lecun1995convolutional}
\bibinfo{author}{LeCun, Y.}, \bibinfo{author}{Bengio, Y.}, et~al.,
  \bibinfo{year}{1995}.
\newblock \bibinfo{title}{Convolutional networks for images, speech, and time
  series}.
\newblock \bibinfo{journal}{The handbook of brain theory and neural networks}
  \bibinfo{volume}{3361}, \bibinfo{pages}{1995}.
\bibitem[{LeCun et~al.(1989)LeCun, Boser, Denker, Henderson, Howard, Hubbard
  and Jackel}]{lecun1989backpropagation}
\bibinfo{author}{LeCun, Y.}, \bibinfo{author}{Boser, B.},
  \bibinfo{author}{Denker, J.S.}, \bibinfo{author}{Henderson, D.},
  \bibinfo{author}{Howard, R.E.}, \bibinfo{author}{Hubbard, W.},
  \bibinfo{author}{Jackel, L.D.}, \bibinfo{year}{1989}.
\newblock \bibinfo{title}{Backpropagation applied to handwritten zip code
  recognition}.
\newblock \bibinfo{journal}{Neural computation} \bibinfo{volume}{1},
  \bibinfo{pages}{541--551}.
\bibitem[{Lu et~al.(2017)Lu, Xiong, Parikh and Socher}]{lu2017knowing}
\bibinfo{author}{Lu, J.}, \bibinfo{author}{Xiong, C.}, \bibinfo{author}{Parikh,
  D.}, \bibinfo{author}{Socher, R.}, \bibinfo{year}{2017}.
\newblock \bibinfo{title}{Knowing when to look: Adaptive attention via a visual
  sentinel for image captioning}, in: \bibinfo{booktitle}{Proceedings of the
  IEEE Conference on Computer Vision and Pattern Recognition (CVPR)},
  p.~\bibinfo{pages}{2}.
\bibitem[{Maunsell and Treue(2006)}]{maunsell2006feature}
\bibinfo{author}{Maunsell, J.H.}, \bibinfo{author}{Treue, S.},
  \bibinfo{year}{2006}.
\newblock \bibinfo{title}{Feature-based attention in visual cortex}.
\newblock \bibinfo{journal}{Trends in neurosciences} \bibinfo{volume}{29},
  \bibinfo{pages}{317--322}.
\bibitem[{Moore and Zirnsak(2017)}]{moore2017neural}
\bibinfo{author}{Moore, T.}, \bibinfo{author}{Zirnsak, M.},
  \bibinfo{year}{2017}.
\newblock \bibinfo{title}{Neural mechanisms of selective visual attention}.
\newblock \bibinfo{journal}{Annual review of psychology} \bibinfo{volume}{68},
  \bibinfo{pages}{47--72}.
\bibitem[{Noudoost et~al.(2010)Noudoost, Chang, Steinmetz and
  Moore}]{noudoost2010top}
\bibinfo{author}{Noudoost, B.}, \bibinfo{author}{Chang, M.H.},
  \bibinfo{author}{Steinmetz, N.A.}, \bibinfo{author}{Moore, T.},
  \bibinfo{year}{2010}.
\newblock \bibinfo{title}{Top-down control of visual attention}.
\newblock \bibinfo{journal}{Current opinion in neurobiology}
  \bibinfo{volume}{20}, \bibinfo{pages}{183--190}.
\bibitem[{Ruder(2017)}]{ruder2017overview}
\bibinfo{author}{Ruder, S.}, \bibinfo{year}{2017}.
\newblock \bibinfo{title}{An overview of multi-task learning in deep neural
  networks}.
\newblock \bibinfo{journal}{arXiv preprint arXiv:1706.05098} .
\bibitem[{Simonyan and Zisserman(2014)}]{simonyan2014very}
\bibinfo{author}{Simonyan, K.}, \bibinfo{author}{Zisserman, A.},
  \bibinfo{year}{2014}.
\newblock \bibinfo{title}{Very deep convolutional networks for large-scale
  image recognition}.
\newblock \bibinfo{journal}{arXiv preprint arXiv:1409.1556} .
\bibitem[{Summerfield and Egner(2009)}]{summerfield2009expectation}
\bibinfo{author}{Summerfield, C.}, \bibinfo{author}{Egner, T.},
  \bibinfo{year}{2009}.
\newblock \bibinfo{title}{Expectation (and attention) in visual cognition}.
\newblock \bibinfo{journal}{Trends in cognitive sciences} \bibinfo{volume}{13},
  \bibinfo{pages}{403--409}.
\bibitem[{Szegedy et~al.(2017)Szegedy, Ioffe, Vanhoucke and
  Alemi}]{szegedy2017inception}
\bibinfo{author}{Szegedy, C.}, \bibinfo{author}{Ioffe, S.},
  \bibinfo{author}{Vanhoucke, V.}, \bibinfo{author}{Alemi, A.A.},
  \bibinfo{year}{2017}.
\newblock \bibinfo{title}{Inception-v4, inception-resnet and the impact of
  residual connections on learning.}, in: \bibinfo{booktitle}{AAAI},
  p.~\bibinfo{pages}{12}.
\bibitem[{Szegedy et~al.(2015)Szegedy, Liu, Jia, Sermanet, Reed, Anguelov,
  Erhan, Vanhoucke and Rabinovich}]{szegedy2015going}
\bibinfo{author}{Szegedy, C.}, \bibinfo{author}{Liu, W.}, \bibinfo{author}{Jia,
  Y.}, \bibinfo{author}{Sermanet, P.}, \bibinfo{author}{Reed, S.},
  \bibinfo{author}{Anguelov, D.}, \bibinfo{author}{Erhan, D.},
  \bibinfo{author}{Vanhoucke, V.}, \bibinfo{author}{Rabinovich, A.},
  \bibinfo{year}{2015}.
\newblock \bibinfo{title}{Going deeper with convolutions}, in:
  \bibinfo{booktitle}{Proceedings of the IEEE conference on computer vision and
  pattern recognition}, pp. \bibinfo{pages}{1--9}.
\bibitem[{Wu et~al.(2016)Wu, Ye, Yuexin and Cohen}]{wu2016encode}
\bibinfo{author}{Wu, Z.}, \bibinfo{author}{Ye, Y.}, \bibinfo{author}{Yuexin,
  Y.}, \bibinfo{author}{Cohen, R.S.W.W.}, \bibinfo{year}{2016}.
\newblock \bibinfo{title}{Encode, review, and decode: Reviewer module for
  caption generation}.
\newblock \bibinfo{journal}{arXiv preprint arXiv:1605.07912} .
\bibitem[{Xu et~al.(2015)Xu, Ba, Kiros, Cho, Courville, Salakhudinov, Zemel and
  Bengio}]{xu2015show}
\bibinfo{author}{Xu, K.}, \bibinfo{author}{Ba, J.}, \bibinfo{author}{Kiros,
  R.}, \bibinfo{author}{Cho, K.}, \bibinfo{author}{Courville, A.},
  \bibinfo{author}{Salakhudinov, R.}, \bibinfo{author}{Zemel, R.},
  \bibinfo{author}{Bengio, Y.}, \bibinfo{year}{2015}.
\newblock \bibinfo{title}{Show, attend and tell: Neural image caption
  generation with visual attention}, in: \bibinfo{booktitle}{International
  conference on machine learning}, pp. \bibinfo{pages}{2048--2057}.
\bibitem[{Yantis and Serences(2003)}]{yantis2003cortical}
\bibinfo{author}{Yantis, S.}, \bibinfo{author}{Serences, J.T.},
  \bibinfo{year}{2003}.
\newblock \bibinfo{title}{Cortical mechanisms of space-based and object-based
  attentional control}.
\newblock \bibinfo{journal}{Current opinion in neurobiology}
  \bibinfo{volume}{13}, \bibinfo{pages}{187--193}.
\bibitem[{Yosinski et~al.(2014)Yosinski, Clune, Bengio and
  Lipson}]{yosinski2014transferable}
\bibinfo{author}{Yosinski, J.}, \bibinfo{author}{Clune, J.},
  \bibinfo{author}{Bengio, Y.}, \bibinfo{author}{Lipson, H.},
  \bibinfo{year}{2014}.
\newblock \bibinfo{title}{How transferable are features in deep neural
  networks?}, in: \bibinfo{booktitle}{Advances in neural information processing
  systems}, pp. \bibinfo{pages}{3320--3328}.
\bibitem[{Zhang et~al.(2017)Zhang, Zhou, Lin and Sun}]{zhang1707shufflenet}
\bibinfo{author}{Zhang, X.}, \bibinfo{author}{Zhou, X.}, \bibinfo{author}{Lin,
  M.}, \bibinfo{author}{Sun, J.}, \bibinfo{year}{2017}.
\newblock \bibinfo{title}{Shufflenet: An extremely efficient convolutional
  neural network for mobile devices. arxiv 2017}.
\newblock \bibinfo{journal}{arXiv preprint arXiv:1707.01083} .

\end{thebibliography}
\end{document}